\documentclass[runningheads]{llncs}
\usepackage[T1]{fontenc}
\usepackage{graphicx}
\usepackage{hyperref}
\usepackage{amsmath}
\usepackage{multirow}

\usepackage{algorithm}
\usepackage{algpseudocode}

\begin{document}
\title{On the Effectiveness of Heterogeneous Ensemble Methods for Re-identification\thanks{This work was supported by the Lamarr-Institute for ML and AI, the Research Center Trustworthy Data Science and Security, the Federal Ministry of Education and Research of Germany and the German federal state of NRW. The Linux HPC cluster at TU Dortmund University, a project of the German Research Foundation, provided the computing power.}}

\titlerunning{Ensemble Methods for Re-identification}

\author{Simon Klüttermann\inst{1, 2}\orcidID{0000-0001-9698-4339} \and Jérôme Rutinowski\inst{1, 2}\orcidID{0000-0001-6907-9296} \and Anh Nguyen\inst{1} \and
Britta Grimme\inst{1}\orcidID{0009-0008-2282-5130} \and 
Moritz Roidl\inst{1}\orcidID{0000-0001-7551-9163} \and
Emmanuel Müller\inst{1, 2, 3}\orcidID{0000-0002-5409-6875}}

\authorrunning{S. Klüttermann et al.}

\institute{TU Dortmund University, Dortmund, Germany \and
Lamarr Institute for Machine Learning and Artificial Intelligence, Germany \and
Research Center Trustworthy Data Science and Security, Germany\\
\email{simon.kluettermann@cs.tu-dortmunde.de}}

\maketitle            

\begin{abstract}
In this contribution, we introduce a novel ensemble method for the re-identification of industrial entities, using images of chipwood pallets and galvanized metal plates as dataset examples. 
Our algorithms replace commonly used, complex siamese neural networks with an ensemble of simplified, rudimentary models, providing wider applicability, especially in hardware-restricted scenarios. 
Each ensemble sub-model uses different types of extracted features of the given data as its input, allowing for the creation of effective ensembles in a fraction of the training duration needed for more complex state-of-the-art models.
We reach state-of-the-art performance at our task, with a Rank-1 accuracy of over 77\% and a Rank-10 accuracy of over 99\%, and introduce five distinct feature extraction approaches, and study their combination using different ensemble methods.

\end{abstract}

\section{Introduction}

Even though the use of deep learning models has become more prevalent for numerous applications, such as industrial computer vision tasks, more shallow approaches remain advantageous in their own right.
This is because the training process of deep learning models requires larger amounts of time and data than their shallower counterparts, and allow less control over the functions learned.
Shallow models, however, tend to yield a lower performance in their respective tasks, mainly because of their lower complexity.
To reduce the overhead of training complex, monolithic deep learning models and to circumvent the poorer performance of shallow models, ensemble learning can be used.
Ensemble models can combine the strengths of multiple simpler models by retaining their training simplicity while improving results \cite{hastie2009ensemble}.
The combination of such models can be realized in many different ways\cite{bagging,boosting}, and the ensembles can also partially or entirely consist of deep learning models\cite{GANAIE2022105151}. 
But while ensembles are a common and straightforward method in supervised machine learning, adapting them to re-identification tasks provides its own challenges.

In industrial settings, providing effective and efficient results is vital: Errors can imply the endangerment of human life or considerable risks and costs.
However, at the same time, the methods used need to be able to work in real-time and often even in hardware-restricted settings, to be economically viable.
Thus ensembles can provide a very practical solution for many applications, in contrast with demanding deep learning solutions \cite{GANAIE2022105151}.

One such industrial application is the re-identification of logistical entities, i.e., inanimate objects used in the context of warehousing.
A prominent logistical entity is the Euro-pallet, which is a widely used type of load carrier, with hundreds of millions of them in constant circulation\cite{deviatkin2020carbon}.
Despite their industrial relevance, Euro-pallets, like other types of standardized pallets, are not equipped with inherent identifiers (e.g., an ID or a barcode) and are instead only identifiable as part of a cluster by, e.g., their place of assembly\cite{din_standards_committee_packaging_pallet_2004}.
This, in turn, does not lead to the individual identification of the pallet but rather its classification (i.e., knowing where a pallet was produced but not making it distinguishable from other pallets produced at the same site).
Due to these limitations, Euro-pallets are usually only identified through documentation such as waybills.

Identifying individual pallets would provide the industry with further knowledge about operational processes, permitting their analysis and improvement \cite{kramer2022blockchain,rutinowski2022potential}, beyond general detection and classification.
This provides users with information on individual pallets and the goods they are carrying in a dispatch network of pallets, permitting process optimization that would otherwise be impossible to realize.
While there have been some suggestions to use artificial markers (e.g., QR-codes), these provide their own challenges through the high number of pallet blocks in circulation and the effects of wear and tear. 
Thus instead, we are trying to recognize the pallet blocks themselves.

A first attempt to solve this challenge, based on the inherent visual characteristics of Euro-pallets, was presented in \cite{rutinowski2021}.
In this contribution, the authors proved the feasibility of identifying the chipwood pallet blocks used in Euro-pallets based on their unique surface structure. 
This work was further expanded upon in \cite{rutinowski2022deep}, in which a larger batch of images, recorded in the industry was successfully used for re-identification.
However, in both contributions, sophisticated deep learning models were employed.
In this contribution, we will insted try to tackle the same challenge by employing ensemble models, in order to provide efficient predictions that still reach a comparable accuracy.

Our code is publicly available at:  \href{https://github.com/psorus/PalletEns}{https://github.com/psorus/PalletEns}

\section{Related Work} \label{sec:rw}
We define re-identification as the attempt to retrieve a previously recorded subject of interest over a network of cameras \cite{ye2021deep}.
Thus, the predictions of a re-identification model are meant to match input data from a query dataset $\mathcal{Q}$ with a given gallery dataset $\mathcal{G}$.
For this purpose, the recorded subjects are assigned a descriptor during their first recording (generated, e.g, by feature extraction).
Based on this descriptor, the similarity between subjects is compared to one another, and the subjects perceived as being the most similar are predicted to be the same \cite{metric1,metric2}.
For this, a list of subjects, ranked by their similarity to the query data ($\mathcal{Q}$), is generated from the gallery dataset ($\mathcal{G}$). 
Prior to the matching task, training occurs using the training dataset ($\mathcal{T}$). 
(These datasets are distinct ($\mathcal{Q}~\cap~\mathcal{G}=\mathcal{Q}~\cap~\mathcal{T}=\mathcal{G}~\cap~\mathcal{T}=\emptyset$), in order to prevent information leakage and to limit overfitting.)
After training, the task of re-identification can then formally be described as matching an image $x_i \in \mathcal{Q}$ of a subject $i$ to an image $y_j \in \mathcal{G}$ of a subject $j$ where $i=j$ \cite{rutinowski2022applicability}.

\subsection{Re-identification Use Cases}
The most common use for re-identification is pedestrian surveillance \cite{islam2023deep,ming2022deep,ye2021deep}, notably in the United Kingdom and the People's Republic of China.
Another application is vehicle surveillance \cite{KHAN201950,wei2018vp}, which is analogous to pedestrian surveillance in its motivation.
Even animals have been shown to be re-identifiable using their inherent visual features \cite{article,ani13050801}.
A more industry-related application is the re-identification of materials \cite{hermanson2011brief,klar1995identifizierung,graph_reident,rutinowski2022deep,rutinowski2022potential,rutinowski2021}.

While different kinds of visual identification criteria can be used for the re-identification of pedestrians, such as their face \cite{s20020342} or their gait \cite{nambiar2019gait} and pedestrians can easily be segmented (e.g., into head, torso, legs), this does not apply to inanimate subjects, such as industrial entities.
Still, first approaches to re-identification of industrial entities do exist \cite{graph_reident,rutinowski2022deep,rutinowski2021,schwenzfeier2023generating}, offering the possibility to identify Euro-pallets by exploiting their unique surface patterns.
In these contributions, PCB\_P4 (Part-based Convolutional Baseline) \cite{sun2018beyond} and graph-based approaches have been deployed.
The results of these studies demonstrate that the surface structure of chipwood and solid wood can be used for the re-identification of Euro-pallets. The datasets used for these works are publicly available
\cite{rutinowski_jerome_2021_6353714,rutinowski_jerome_2022_6358607}.
These works, however, do not employ ensemble methods and are only applied to a single logistical entity, namely Euro-pallets.

\subsection{Re-identification Approaches} 

An example for a re-identification approach is metric learning, which maps the subject's similarity as feature embeddings, describing similarity as distances between two points in this embedding space \cite{ming2022deep,Paisitkriangkrai_2015_CVPR}.
As loss functions, classification loss, constrastive loss and, most prominently, triplet loss are often used to minimize intra-class distances and maximize inter-class distances \cite{islam2023deep,ming2022deep,class_loss}.

As to evaluate re-identification results, the most common metric used is ranked accuracy, also known as top-k accuracy.
This metric describes whether the query image is retrieved in the first $k$ matches of the gallery data \cite{ming2022deep,ye2021deep}. 
This adds explainability and transparency to the model's predictions, providing insight into the severity of a wrong retrieval.
This insight can be especially useful, as there is a practical difference for many industrial use cases, between a prediction that is off by one rank or multiple ranks.

\subsection{Ensembles Usage for Re-identification}

Using ensemble models is a common practice in machine learning, allowing the combination of different predictors to achieve a better performance \cite{ensemble-survey}. 
There are many methods, ranging from the simple averaging of multiple models (bagging) \cite{bagging} to training new models on the prediction of the ensemble sub-models (stacking) \cite{stackingad}. 
However, there are surprisingly few applications to re-identification, as this requires us to adapt concepts like bagging and stacking. 
We want to summarize the existing approaches in this chapter.

A common approach to re-identification focuses on extracting general features representing aspects of a sample and using their resemblance as an indicator of similarity.
This invites the use of ensembles to combine different features.
A previous publication \cite{earlyreidensemble} employs ensembles by combining different hand-crafted color descriptors, which allows the extraction of information from more than one feature. 
Recent approaches \cite{featreidens4,featreidens3,featreidens2} apply more advanced neural networks to find features using pre-trained models or to extract them directly.

The drawback of feature extraction is that many datasets contain a large amount of superfluous features, which tend to be included in well-studied, commonly extracted features. 
Additionally, there has not been enough research conducted yet to find specialized features for pallet re-identification and a solution emerging from such research would only be applicable to one use case or dataset, thus not leading to a general solution.
Siamese neural networks\cite{siamese}, however, are able to effectively extract useful features automatically, providing a more generalizable and adaptable solution.
Other approaches use ensembles for metric learning \cite{metricreidens} or to better handle multiple data modalities \cite{enscrossmodalityreid}.

However, to the best of our knowledge, ensembles have hardly been studied for siamese neural networks.
Worth mentioning is \cite{siameseEnsNLP}, in which ensembles are used to enhance contrastive learning in the context of natural language processing.
While this ensemble method shares similarities with our approach, the application is fundamentally different, which limits the transferability of results. 

\section{Methodology} \label{sec:methodology}

This section aims to provide the reader with insights into our methodology.
The section delves deeper into re-identification task description for the given use case.
It also describes the re-identification methods that are used for this task as well as the ensembles that are comprised of the former.

\begin{figure}
    \centering
    
    \begin{minipage}{0.655\textwidth}
        \centering
        \includegraphics[width=\textwidth]{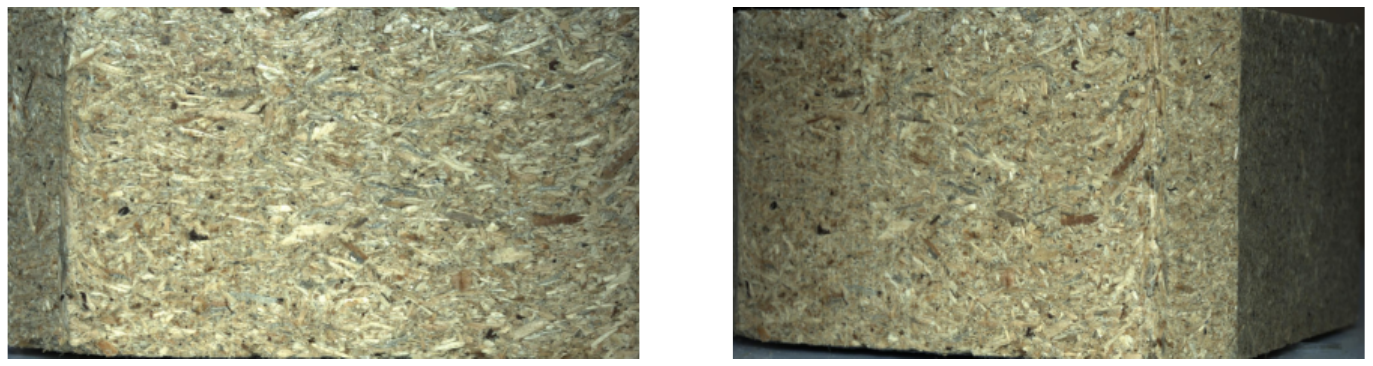}
(a)
            \end{minipage}\hfill
            
    \begin{minipage}{0.65\textwidth}
    \vspace{8pt}
        \centering
        \includegraphics[width=1\textwidth]{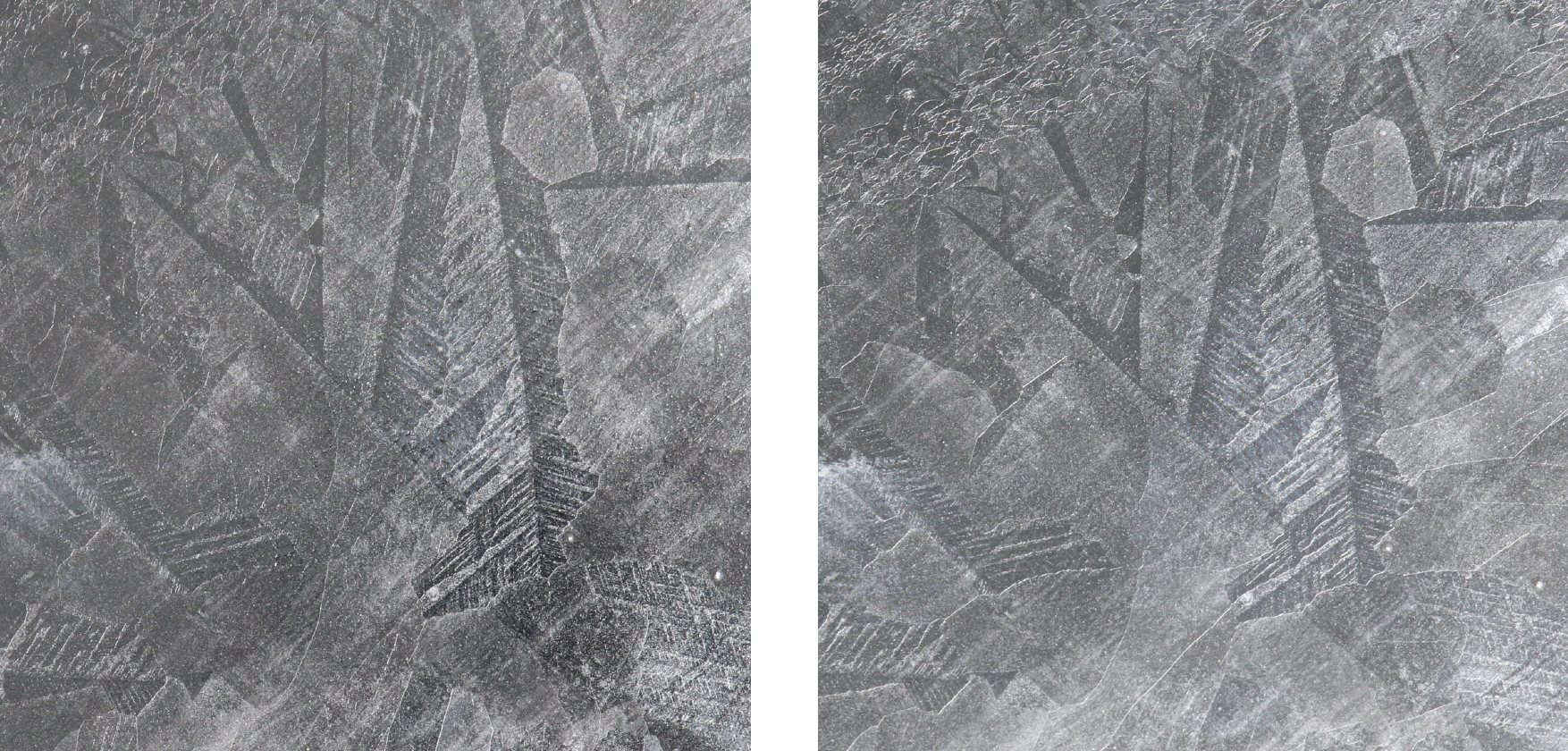}
(b)
    \end{minipage}
    
    \caption{Excerpts of the used datasets. a) pallet block dataset  \cite{rutinowski_jerome_2021_6353714}, b) galvanized metal dataset \cite{rutinowski_jerome_2022_7386956}}
    \label{fig:quick}

\end{figure}

As a first step, we describe the herein used datasets (see Fig. \ref{fig:quick}).
The first dataset \cite{rutinowski_jerome_2021_6353714},  containing images of wooden pallet blocks, is an open-source dataset that is freely available online, providing insights into the dataset creation process and recording setup.
It comprises $5,020$ RGB images of pristine, unbranded EPAL pallet blocks.
The images in the dataset have been recorded using two different cameras and five different perspectives: central, left-hand and right-hand side rotation, and left-hand and right-hand side shift.
Half of those images ($2,510$), taken by camera 1 in different rotations, are used for the experiments in this contribution.
Other parts of the dataset, as well as its entirety, could also be used for the experiments.
With $10$ images recorded per pallet block, the dataset contains images of a total of $502$ different pallet blocks.

Additionally, we use a second dataset \cite{rutinowski_jerome_2022_7386956}, which contains $5,088$ images of galvanized metal plates.
Galvanized metal, like wood, is a commonly used material (e.g., for pallet cages or metal crates) and also possesses unique surface structure that makes it identifiable. 
The plates used for this dataset are recorded using ambient light and photo light, and both at a 90° and a 75° angle.
For our experiments, we used the $2,544$ images of the dataset recorded with photo lighting under 75° and 90° angles.

\subsection{Task Description}

The task of re-identification is often solved by learning a transformation $f$ of a dataset such that the distances between the representations of the same entity are minimized and the distances between different ones are maximized. This goal can be formalized as follows (with $x^i$ being a sample following concept $i$):
\begin{equation}
\|f(x_a^i)-f(x_b^j)\|_2=\begin{cases}
\text{small if } i = j\\
\text{large if } i \neq j\\
\end{cases}
\forall x,y
\end{equation}

This task can be achieved by minimizing a suitable loss function, e.g., triplet loss \cite{triplet}:    

\begin{align}
\label{eqn:triplet}
L_{triplet} = 
\max (&0, \|f(x^i_a)-f(x^i_b)\|_2 \\&- \|f(x^i_a)-f(x^j_c)\|_2 +\alpha)\;,\; 
i \neq j\nonumber
\end{align}

where $x^i_a$,$x^i_b$ and $x^j_c$ represent three samples of two different subjects $i$ and $j$ and $\alpha$ is a hyperparameter. 
This loss value is averaged over a large number of different triplets.

Our goal in this contribution is to study the case in which we have not only a single transformation $f$ but also a second transformation $g$. Therefore, we define and investigate an ensemble $e$ including both transformations, with an ensemble creation transformation $T$ creating an ensemble $e(x)=T(f(x)\oplus g(x))$.

\subsection{Re-identification Methods} \label{sec:reidmethods}

In order to study ensembles, we require sub-models. 
We chose heterogeneous sub-models, as this allows using pre-existing knowledge about our dataset. 
Additionally, this is a faster approach, as a lower number of sub-models is required.
To the best of our knowledge, this is the first application of heterogeneous ensembling for re-identification.
However, this also means that we might need to change the sub-models depending on the dataset, in order to maximize the ensemble's prediction accuracy.
Here, we suggest five siamese models for each dataset, all based on different approaches:

\textbf{Image-based re-identification:}
The most direct approach for training a siamese neural network on image-based data is to use convolutional neural networks.
The benefit of this approach is its ability to extract information available in the entire image.
However, this also represents the most time consuming approach, and is the most susceptible to overfitting.
Further information on the training process of this model can be found in the supplementary material.

\textbf{Graph-based re-identification:}
For our graph-based approach, we manually extract the most essential information before training a siamese neural network on it.
For this, we use the approach implemented in \cite{graph_reident}, where anomalous parts of an image are considered nodes in a graph.
Notice that our performance differs from the one stated in \cite{graph_reident}, as we use a different fraction of query and gallery data. Alas, this approach is specialized to the pallet block dataset, and our attempts to generalize it to the metal dataset were unsuccessful. 
Thus, we only use it on the first dataset.

\textbf{Linear quantile re-identification:}
To compensate for the lack of applicability of the graph based approach to the metal dataset, we developed a different model, exploiting the perspective-wise shearing in the galvanized metal dataset. 
For this purpose, we split the images into pixel columns and use the $0.2$, $0.5$ and $0.8$ quantile of each pixel value in this column as an input for a siamese neural network.

\textbf{Brightness, average color and color variance based re-identification:}
The remaining three approaches split a given image into sub-images of size \mbox{$16\times16$ px} each. 
We iterate over all sub-images of an image (with $50\%$ overlap) and calculate $1$ -- $3$ values for each, resulting in a representation with $768$/$2304$ dimensions.
For the \emph{brightness}-based approach, we calculate the mean of each sub-image, while for the \emph{average color}-based approach, we calculate a mean for each color and sub-image. 
Finally, for the \emph{color variance}-based approach, we use the standard deviation of each color throughout the sub-image. 
The values for each sub-image are concatenated into a vector and used to train a siamese neural network.
Each network uses three dense layers with $100$ nodes each to generate a \mbox{$50$-dimensional} representation. 
While these approaches ignore much of the available information given in an image, they are also straightforward and time-efficient, reducing the required training time by multiple orders of magnitude.

\subsection{Ensembles} \label{sec:ensemblemethods}

In this contribution, we study four different transformations $T$ that create ensembles.
While ensembles are used in many cases (see Section \mbox{\ref{sec:rw}}), they are only rarely used for siamese network-based re-identification. 
We therefore aim to use ensemble models and study their applicability for the task at hand, since an ensemble usage would allow us to benefit from the combination of rudimentary methods instead of relying on a single, complex model, providing time-efficient and computationally less demanding results.

The simplest ensemble transformation presented in this work, the \emph{Concatenation} transformation, normalizes each ensemble component using the mean and the standard deviation (\emph{std}) for all considered samples and concatenates them. It can be seen as an ensemble method following the concept of bagging \cite{bagging}.

\begin{equation}
   T_{Concat}((x_0,..., x_n))= (z_0,...,z_n)\;\text{with }\;z_j^i =  \frac{x_j^i - \bar{x_j}}{s_{x_j}}
\end{equation}
Here $x_j^i$ is denoting the \emph{j}-th component of the representation of sample $x^i$. 
Meanwhile, the \emph{Neural Network Triplet} transformation is a stacking method \cite{stackingsurvey}, which uses a neural network to represent $T_{NN triplet}$ and to train it by minimizing Eqn. \ref{eqn:triplet} on the training set $\mathcal{T}$.
The \emph{Weighted Triplet} instead simply uses weighing factors:
\begin{equation}
    T_{W triplet}(f(x) \oplus g(x))=\alpha_f f(x) \oplus \alpha_g g(x)
\end{equation}
These are found by minimizing Eqn. \ref{eqn:triplet} using gradient descent.
Subsequently, the \emph{Weighted Accuracy} transformation tries to maximize the probability that the closest sample $y^j \in \mathcal{G}$ to a sample of the test set $x^i \in \mathcal{Q}$ depicts the same subject $i=j$.
Since this probability is not continuous, gradient descent cannot be used for this task.
Instead, the probability is optimized using the \textit{flaml} library \cite{flaml}, which is designed to optimize non-continuous hyperparameter optimization problems, and thus does not rely on gradient descent.
Additionally, a fifth ensemble, called the \emph{Majority Vote} is created using a different approach to ensembling.
For this ensemble, multiple sub-models do not calculate the distance to each gallery sample $\in \mathcal{G}$, but simply order the gallery samples based on their distance to the query sample.
We choose the most common index of this order through all sub-models as an indicator of similarity between samples.

\section{Experiments}\label{sec:exp} 

To test our novel approaches and to evaluate which ensemble method yields the highest performance, we train a siamese neural network on each of the data pre-processing steps described in Section \ref{sec:reidmethods}.
This creates five different representations of each image, in which samples representing the same object are closer to each other than to different ones.
While some pre-processing steps retain more information, others are able to express the information they contain in a more efficient manner.

To provide more meaningful results and to estimate prediction uncertainties, we employ cross-validation. 
We split our dataset into five different groups on the pallet block dataset and six different groups on the metal dataset. 
One of these folds is chosen as a \emph{query} and \emph{gallery} set, in which for each pallet block, one image is selected for the \emph{query} set, while the rest become part of the \emph{gallery} set. 
One fold is not used but held out for a possible future novelty detection task, while the remaining folds are used as a \emph{training} set.

\subsection{Individual Results}

\begin{figure}[hb!]
    \centering
    \begin{minipage}{0.49\textwidth}
        \centering
        \includegraphics[width=\textwidth]{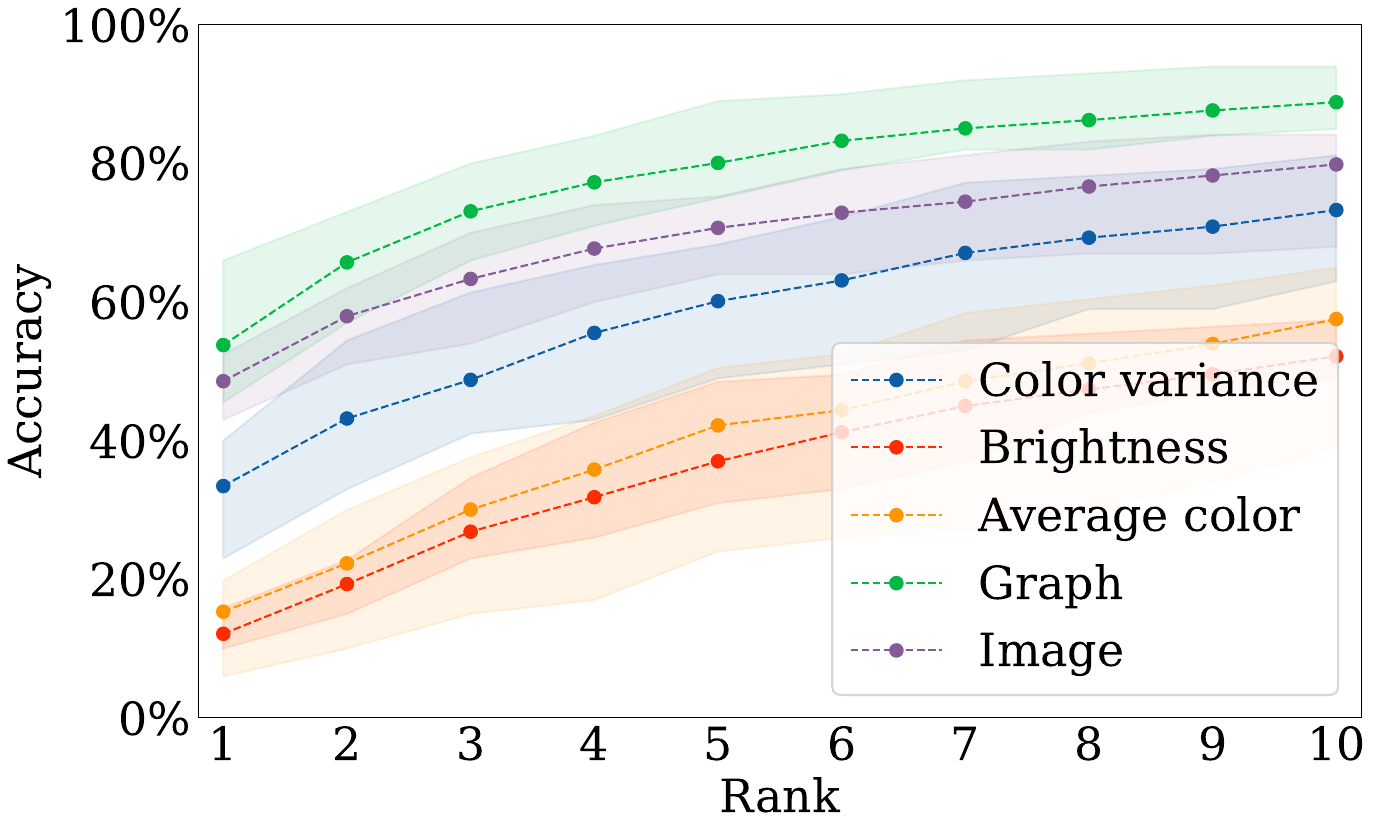}
        (a)
    \end{minipage}\hfill
    \vspace{8pt}
    \begin{minipage}{0.49\textwidth}
        \centering
        \includegraphics[width=\textwidth]{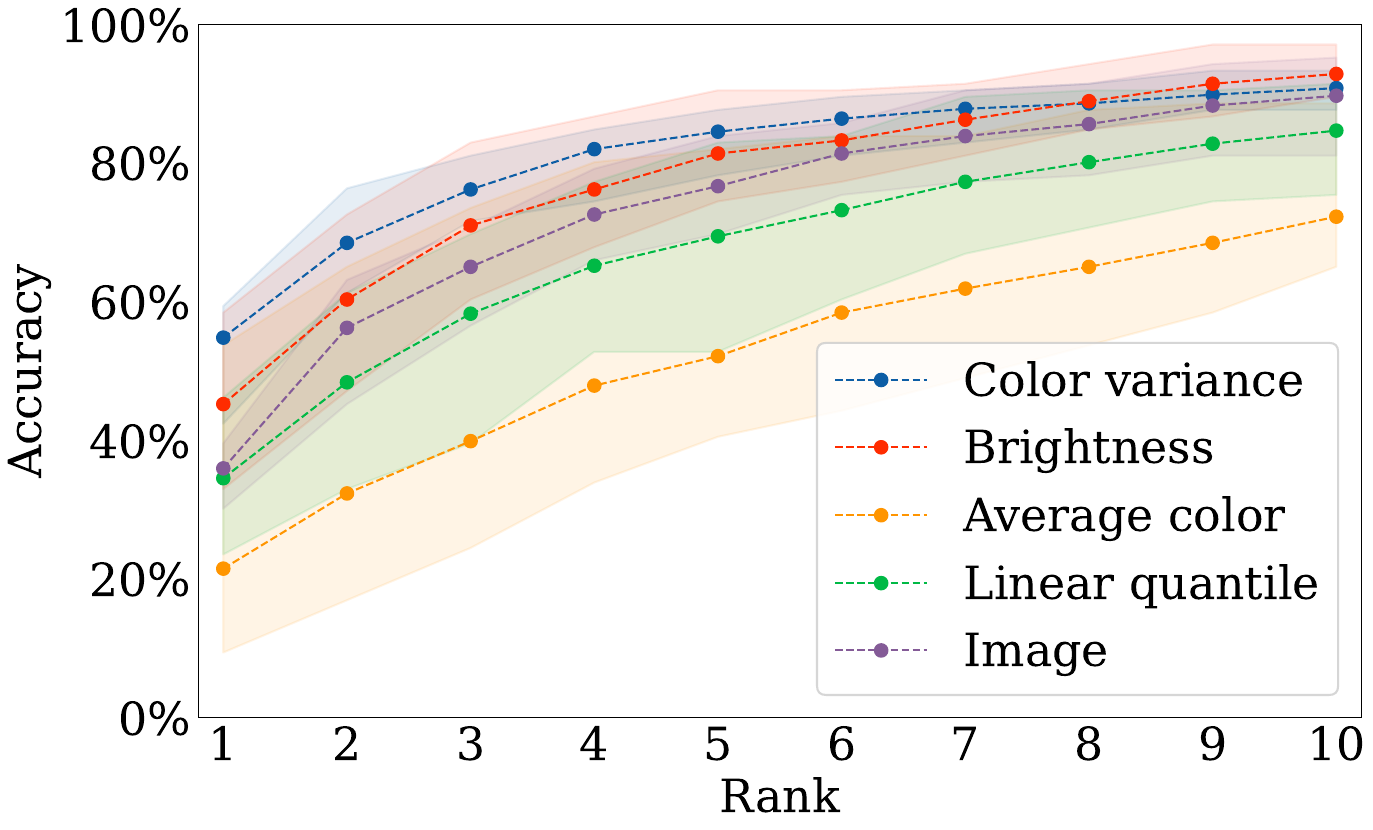} 
    (b)
    \end{minipage}
    \caption{Rank-k accuracy for the ensemble sub-models. The highlighted areas represent the highest and lowest scores per individual fold. a) pallet block dataset, b) metal dataset}
    \label{fig:individual}

\end{figure}

First, we evaluate the individual sub-models and plot the resulting \mbox{Rank-1} to Rank-10 accuracy, as shown in Fig. \ref{fig:individual}.

The two methods averaging color values (\emph{brightness} and \emph{average color}) provide subpar results on the pallet block dataset, with their overall performance reaching a Rank-1 accuracy of \mbox{$<20\%$}.
An entirely different effect can be observed for the metal dataset, with the simpler \emph{brightness}-based method outperforming a method with image-wide information, demonstrating the effect that a careful selection of sub-models can provide.

Still, in both cases, using the \emph{color variance}-based model provides a higher accuracy.
We assume that this is due to high variances often representing unusual structures that can be used for re-identification, while in an average-based method these structures are averaged out. 
Even though the performance of these simple methods are not always competitive, they are an order of magnitude faster to train than other methods (i.e., $\sim8s$ compared to about $20min$ for the Inception model) and thus can be valuable ensemble sub-models.

Notice that on both datasets, the best performance was not achieved using the highest number of available features.
This implies the notion that performance can further be improved using an ensemble method.

\subsection{Ensemble Methods}

\begin{figure}[h!]
    \centering
    \begin{minipage}{0.49\textwidth}
        \centering
        \includegraphics[width=\textwidth]{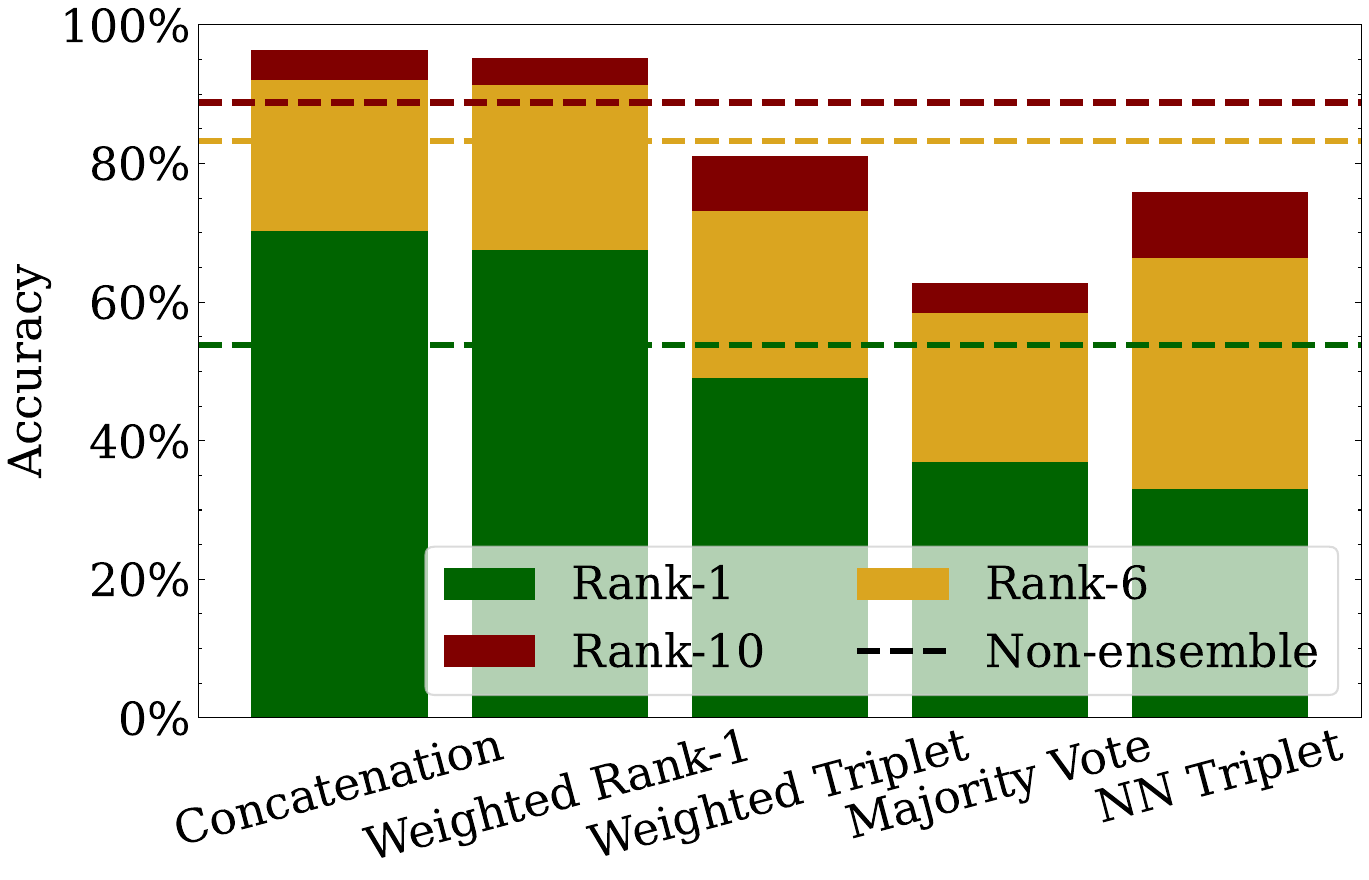} 
        (a)
    \end{minipage}\hfill
    \vspace{8pt}
    \begin{minipage}{0.49\textwidth}
        \centering
        \includegraphics[width=\textwidth]{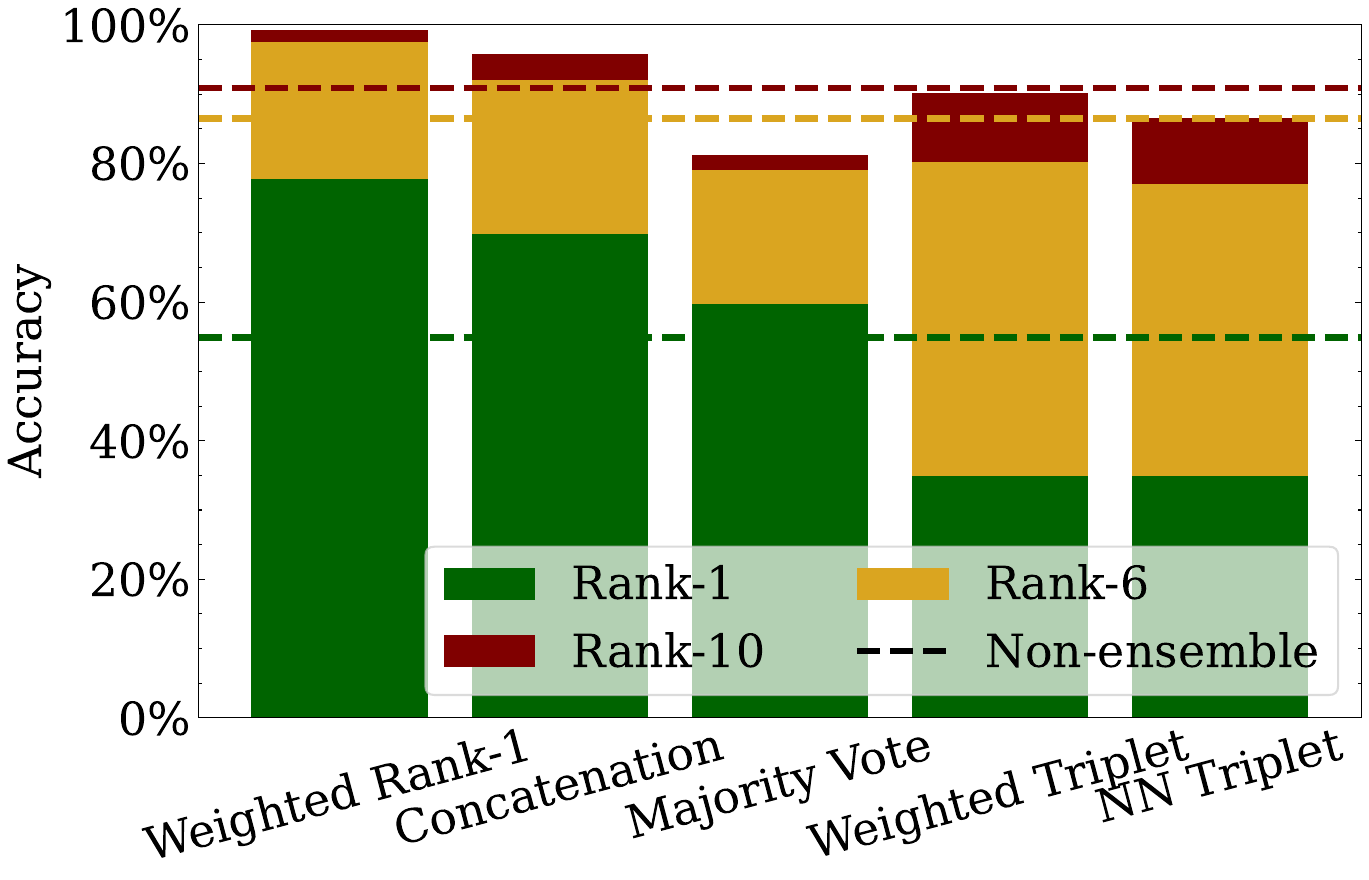} 
(b)
    \end{minipage}
        \caption{Rank-1, Rank-6, and Rank-10 accuracy for the five different ensemble approaches compared to our most effective non-ensemble model (the \emph{graph}-based model). a) pallet block dataset, b) metal dataset}
      \label{fig:ensembles}

\end{figure}

In this section, the performance of the ensemble models is studied. 
Fig. \ref{fig:ensembles} show the performance of the five different ensemble models discussed in Section \ref{sec:ensemblemethods}.

The ensemble models provide much higher performances than the individual models. 
Given the results in Fig. \ref{fig:ensembles}, the ensembling approach yielding the highest accuracy on the pallet block dataset seems to be the concatenation, which allows us to improve the Rank-1 accuracy from $54\%$ to over $70\%$, compared to the graph-based sub-model.
On the metal dataset, the concatenation also performs well, but here, learning weights seems valuable, increasing the Rank-1 accuracy from $55\%$ to almost $78\%$, compared to the best sub-model.
It is quite counter-intuitive that the simplest (and unsupervised) approach to ensembling also yields one of the highest performances.

We studied three ways of weighing individual sub-models and expected the weighing approaches to improve results, given that Fig. \ref{fig:individual} showed significant differences of the sub-models' performances. 
We would also have expected similarly functioning approaches (like our \emph{brightness} and \emph{average color}-based approaches) to potentially cancel one another out. 
However, while weighing seems to increase the resulting performance in some cases, it seems not always to be an effective solution.
One reason for this phenomenon might be the weighing approaches' susceptibility to overfitting.

The lowest (Rank-1) accuracy is obtained by the neural network approach, for which we use another neural network trained on the output of each siamese neural network to minimize the triplet loss (Eqn. \ref{eqn:triplet}) and generate a new common representation. 
This matches our expectation since the neural network also contains the highest amount of parameters and thus has the highest potential of overfitting.
Slightly better results are provided when a weighting factor is added and optimized to the output of each sub-model (\emph{Weighted Triplet}).
Still, the performance is inferior to that of the best-performing sub-model. 
We expect this to be a result of a local minimum of the optimization task because when changing the function to be optimized from a triplet loss to a Rank-1 accuracy on the training set (\emph{Weighted Accuracy}) the performance can be improved, up to a level comparable to the performance of the \emph{Concatenation} approach. 
In any case, this implies that there remains untapped potential for the application of more sophisticated ensemble methods.

We tried another approach, in which the median prediction of each sub-model is chosen (\emph{Majority Vote}). Still, this method does not perform well either.
Through this ensemble method usage, we effectively perform an abstract segmentation of the data, analogous to the aforementioned pedestrian segmentation approaches, which however is restricted to this very use case, unlike our approach.
Therefore, while other high-performance methods, especially in pedestrian re-identification are highly specialized for their specific domain of application, our approach yields a high degree of transferability, as we also demonstrate by applying it to two distinct dataset.

\subsection{State of the Art Comparison}

Finally, we compare our method to commonly used approaches (see Tab. \ref{tab1}). 
In addition to the previously introduced models, we compare our models to a method similar to one used for Person Re-identification\cite{islam2023deep,sun2018beyond}. 
The approach here is to use a pre-trained model, to add three Dense layers for the output to match, and to continue training the model. 
The main difference to common person re-identification models, is that we employ the InceptionV3 object detection model \cite{inception}, instead of a model trained to re-identify pedestrians. We chose this model over, e.g., other state-of-the-art models like ResNet50\cite{resnet50} and MobileNetV2\cite{mobilenetv2} as it seems to provide significantly higher performance in our experiments.
Still, this approach is not able to outperform our ensemble methods while requiring significantly more time to train.

\begin{table}[ht]
    \centering
    \caption{Comparison of our method with alternative approaches on the pallet block and metal datasets.}
    \label{tab1}
    \setlength{\tabcolsep}{7pt}
    \begin{tabular}{llrrrrr}
        \hline
        \multicolumn{1}{c}{Dataset} & \multicolumn{1}{c}{Model} & \multicolumn{1}{c}{Rank-1} & \multicolumn{1}{c}{Rank-10} & \multicolumn{1}{c}{Runtime [s]}\\
        \hline
        \multirow{4}{*}{Pallet Block} & Ensemble & $\textbf{0.703} \pm 0.079$ & $\textbf{0.964} \pm 0.020$ &  $287$ \\
         & Graph \cite{graph_reident}    & $0.526 \pm 0.052$ & $0.904 \pm 0.045$ & $50$ \\
         & Image \cite{rutinowski2021}   & $0.486 \pm 0.032$ & $0.799 \pm 0.060$ &  $213$ \\
         & Inception \cite{inception} &  $0.518 \pm 0.05$ & $0.918 \pm 0.014$  &  $1,116$\\
        \hline
        \multirow{4}{*}{Metal} & Ensemble      & $\textbf{0.777} \pm 0.054$  & $\textbf{0.992} \pm 0.007$ & $257$\\
         & Color Variance & $0.549 \pm 0.057$ &  $0.909 \pm 0.022$  & $8$\\
         & Image \cite{rutinowski2021}         & $0.360 \pm 0.033$ & $0.898 \pm 0.047$  & $224$\\
         & Inception \cite{inception} & $0.681 \pm 0.059$ &  $0.951 \pm 0.013$ & $1,283$\\
        \hline
    \end{tabular}
\end{table}

It is also again apparent, that our best individual sub-models are significantly outperformed by the ensemble model.
Focusing on Rank-10 accuracy, the results of the four models in question converge between $79.9$\% and $99.2$\%, with the image-based model consistently providing the lowest re-identification accuracy.
The accuracies obtained using the metal dataset were consistently higher than the ones using the pallet block dataset, implying a potential difference in difficulties between the two re-identification tasks.
Further analysis of our ensemble model is found in the supplementary materials provided.

\section{Conclusion} \label{sec:conclude}
In this contribution, the first study of heterogeneous siamese neural network ensembles for re-identification purposes has been presented.
Using such ensembles, we reached state-of-the-art performance on our datasets.
With our novel approach, we obtain Rank-1 re-identification accuracies of $70\%$ and $77\%$ on the pallet block and metal datasets, respectively.
One of our employed ensembling methods, the \emph{Concatenation} method, a fairly simple approach, performs better than most of the more sophisticated ones evaluated in this work.
In subsequent publications, we aim to study the use of a higher number of possibly even simpler models, while still aiming to maintain reliable re-identification results.
We also show, how our ensemble approach can be applied and specified to different datasets, implying that it might be possible to apply our methods to further contrastive learning tasks.
Since we restricted ourselves to the application for the re-identification of logistical entities, we invite the research community to apply similar approaches to other datasets.
We also invite researchers to apply their approaches to the same datasets, to provide a comparison.
Finally, we believe the further investigation of the explainability and trustworthiness of re-identification methods to be a promising research direction.
For this reason, we expand upon this topic in our supplementary material in great detail.

\bibliographystyle{splncs04}
\bibliography{refs}

\newpage    
\appendix

\section{Supplementary Material -- Trustworthiness}

\subsection{Motivation}
A largely unstudied benefit of ensembles is the increased trustworthiness they can provide to tasks like re-identification. 
Next to the increased performance, ensembles also tend to generalize better to unseen data \cite{ensemblegeneralisation} by allowing sub-model errors to cancel each other out \cite{ensembleerrorcancel}. 

Since we use heterogeneous sub-models, we can also use the difference in errors between sub-models to learn more about our algorithms and dataset.
This can allow us to remove biases introduced by different sub-models to mitigate sensitivities, e.g., to changes in brightness or shearing effects. 
By using simple approaches to re-identification, it could also be extended to provide explanations for why two samples are considered similar to each other.
Visualizing the sub-models results might provide an increase in explainability and trustworthiness:
It allows the resulting matches and mismatches to be studied and to find and alter features that lead the models to make erroneous predictions. 
We believe this to be of great value to users, providing them with models that hold a greater degree of explainability and which can therefore ultimately be considered to be more trustworthy.
Additionally an important part of trustworthiness is the reliability of the results over multiple experiments.
Thus, considering it a measure of trustworthiness, we also study the uncertainty of our results, by employing cross-validation.

\subsection{Results}\label{sec:reliable}

Additionally, we want to study the effect that using ensembles has on increasing the reproducibility, reliability and explainability of our resulting model. 

\begin{figure}[h!]
\centerline{\includegraphics[width=0.8\columnwidth]{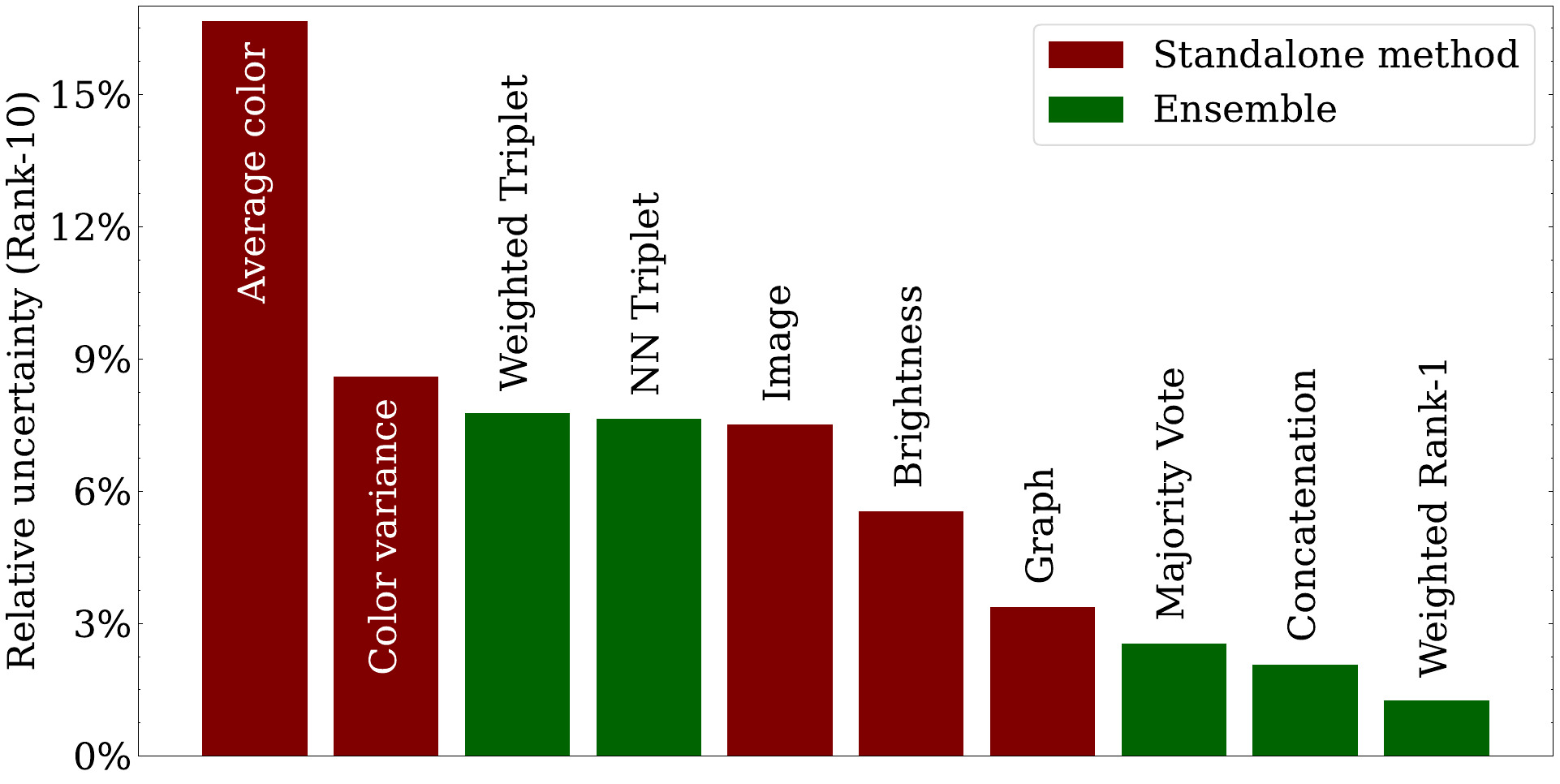}}
\caption{Relative uncertainty (accuracy uncertainty divided by accuracy) of the Rank-10 accuracy for different models.}
\label{fig:reliable}
\end{figure}

First, we present the relative uncertainty of the Rank-10 accuracy of each of our individual and ensemble approaches in Fig. \ref{fig:reliable}.

We choose to present Rank-10 instead of Rank-1 accuracy, as these uncertainties are naturally quite noisy, which is reduced by the higher rank.
Both ensembles that outperform each individual model in terms of accuracy also have the lowest uncertainty.
They reach an uncertainty that is almost an order of magnitude smaller than the \emph{average color} approach and clearly lower than all individual models. 
This increased reliability can be a crucial benefit to the implementation of our method in industrial settings.

\begin{figure}[htbp]
\centerline{\includegraphics[width=0.8\columnwidth]{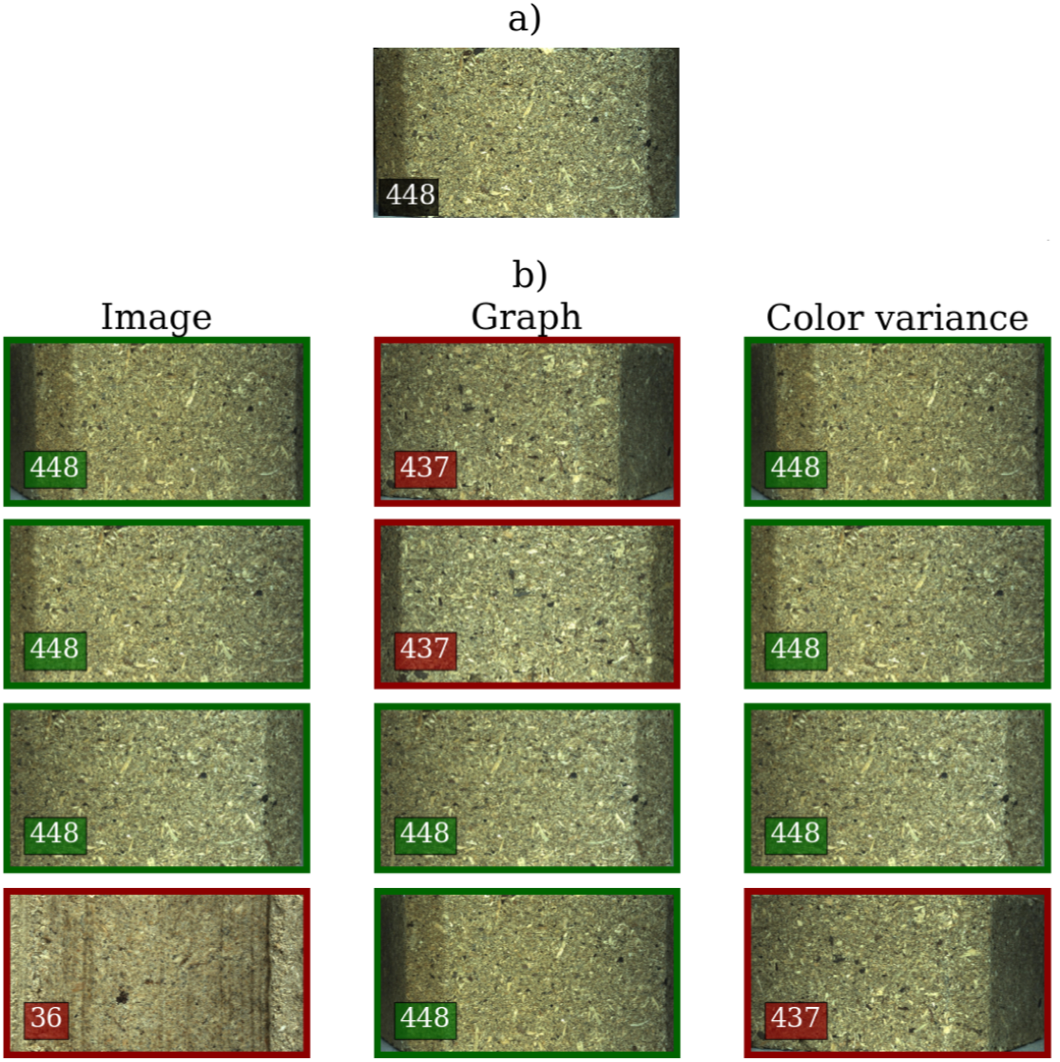}}
\caption{Rank-1 to Rank-4 gallery matches (b) to a query image (a), for three sub-models. The Image ID is indicated in the corner, correct matches are framed in green, incorrect ones in red.}
\label{trust}
\end{figure}

Finally, we present a visualization of our sub-models' ranked results in Fig. \mbox{\ref{trust}}.
This visualization shows how different methods can come to different results, i.e., in this instance the comparison between image-based, graph-based, and color variance-based models that lead to different rankings of the same set of images, making it possible to debug methodological problems.

\section{Supplementary Material -- Additional Analysis}
\subsection{Time Efficiency}
\begin{figure}[htbp]
\centerline{\includegraphics[width=0.8\columnwidth]{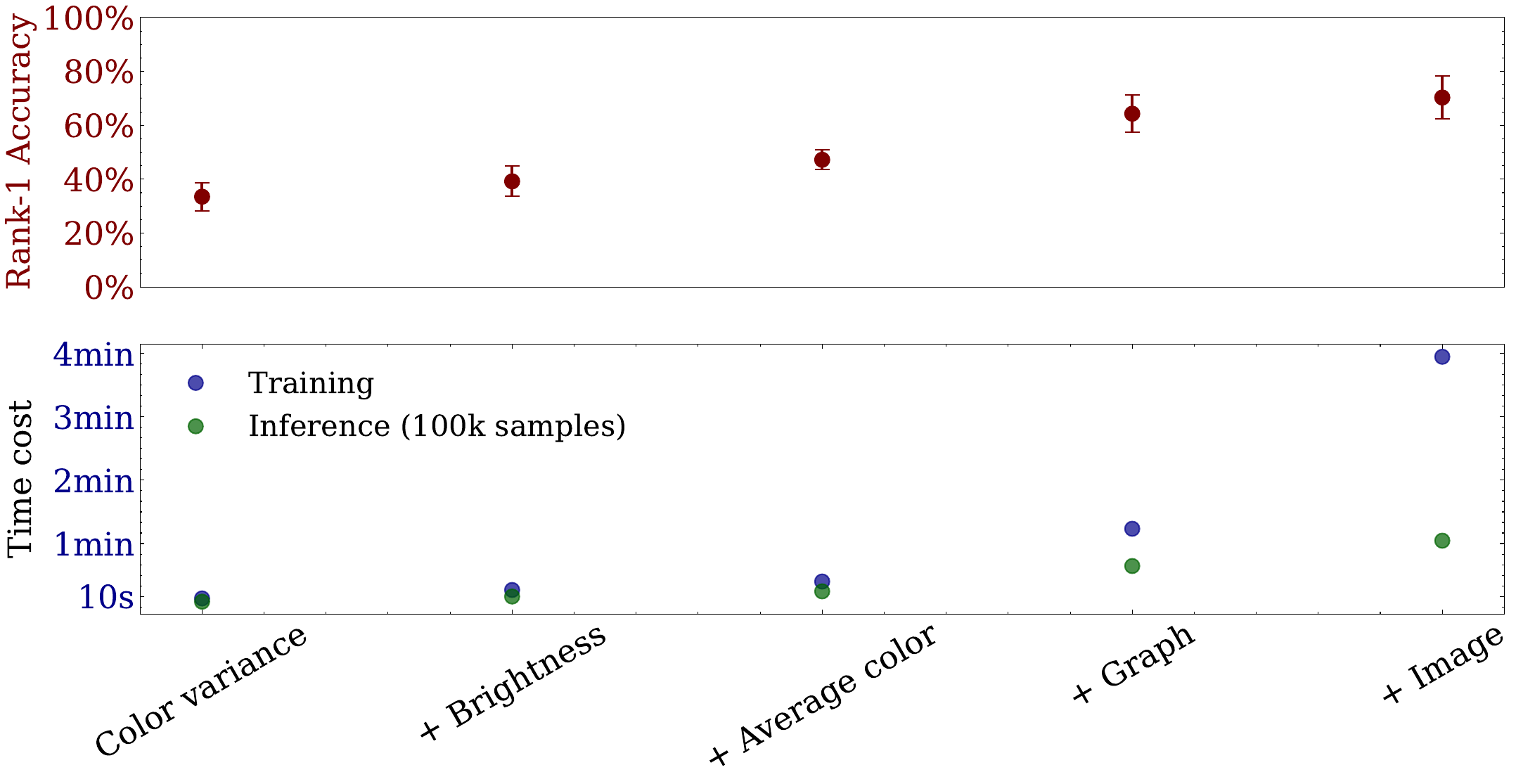}}
\caption{Training duration, prediction cost and Rank-1 accuracy of different ensembles. The sub-models are chosen to minimize the training duration. All our experiments were conducted using an NVIDIA A100 graphics card with 40GB of VRAM.}
\label{fig:justice}
\end{figure}

While each sub-model improves the ensemble performance, some of our models require significantly more time to be trained than others.
This leads to a trade-off between training effort and re-identification performance.
Motivated by this, we display both the Rank-1 accuracy and the training duration of partial ensembles in \mbox{Fig. \ref{fig:justice}}.

An ensemble of all three \emph{color}-based methods can be trained in a few seconds and still performs almost as well as the \emph{graph}-based method with a significantly higher training cost.
This highlights another use case of heterogeneous ensembles: besides increasing the overall performance of a re-identification method it may also contribute to decreasing the time required for achieving comparable performance.
We expect an ensemble of many more simple sub-models to outperform single complex models and be easier to use.

\subsection{Contribution Studies}

\begin{figure}[htbp]
\centerline{\includegraphics[width=0.8\columnwidth]{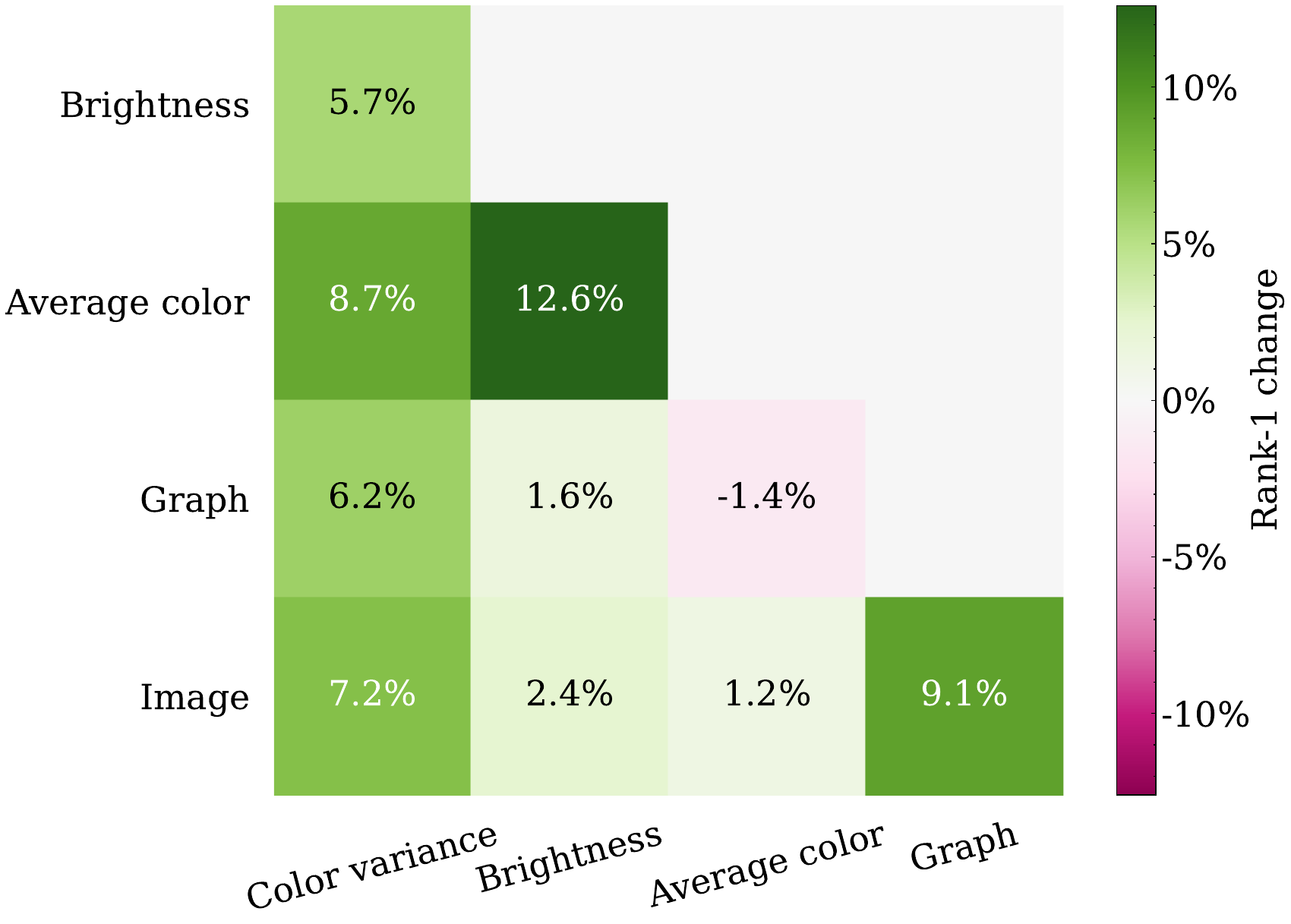}}
\caption{Rank-1 accuracy change for all possible two-model ensembles compared to their best sub-model performance.}
\label{fig:duals}
\end{figure}

To study our ensembles further, we present the Rank-1 accuracy improvement of an ensemble compared to its best individual model for all two-model ensembles in Fig. \ref{fig:duals}.

First, it is apparent that almost all Rank-1 changes are positive. 
Only the combination of a \emph{graph} representation with an \emph{average color} leads to a negative change in Rank-1 accuracy. 
We hypothesize that this might result from the high uncertainty of the \emph{average color} model and the drastic difference in the performance of both models.
Most importantly, the sum of each row and column is positive, implying that adding another model always benefits the ensemble (see also Fig. 2 in the supplementary material). 
This might also explain why we do not need any weighting factors to achieve higher performance; and it implies that even more ensemble sub-models might help to improve the ensemble's performance further.

\subsection{Ablation study on representation sizes}
\begin{figure}[h!]
\centerline{\includegraphics[width=0.8\columnwidth]{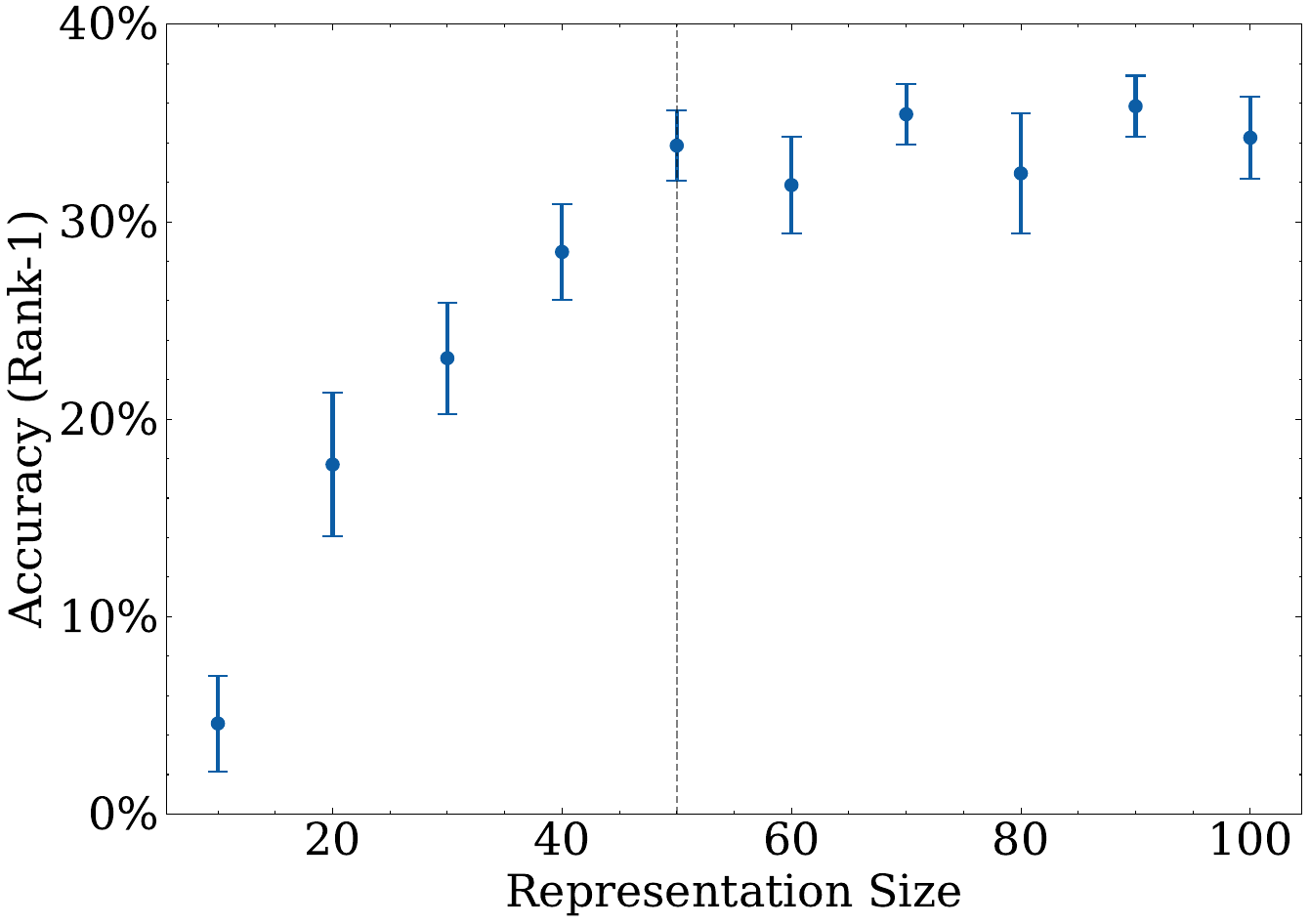}}
\caption{Ablation study: Rank-1 accuracy of the color variance approach, as a function of the representation size used.}
\label{fig:size}
\end{figure}

We tested various representation sizes for the color variance approach, to determine which representation size to choose (in this case, a representation size of $50$). This is shown in Fig. \ref{fig:size}. 

\newpage
Here, the performance seems to plateau from a representation size of $50$ onward, which is why we decided to use it. Similarly we chose the same size for the rest of the sub-models, except for the image-based network. Here, we chose a higher number ($100$), as the higher number of input features might relate to a higher number of output features.

\subsection{Singular sub-model impact}
\begin{figure}[htbp]
\centerline{\includegraphics[width=0.8\columnwidth]{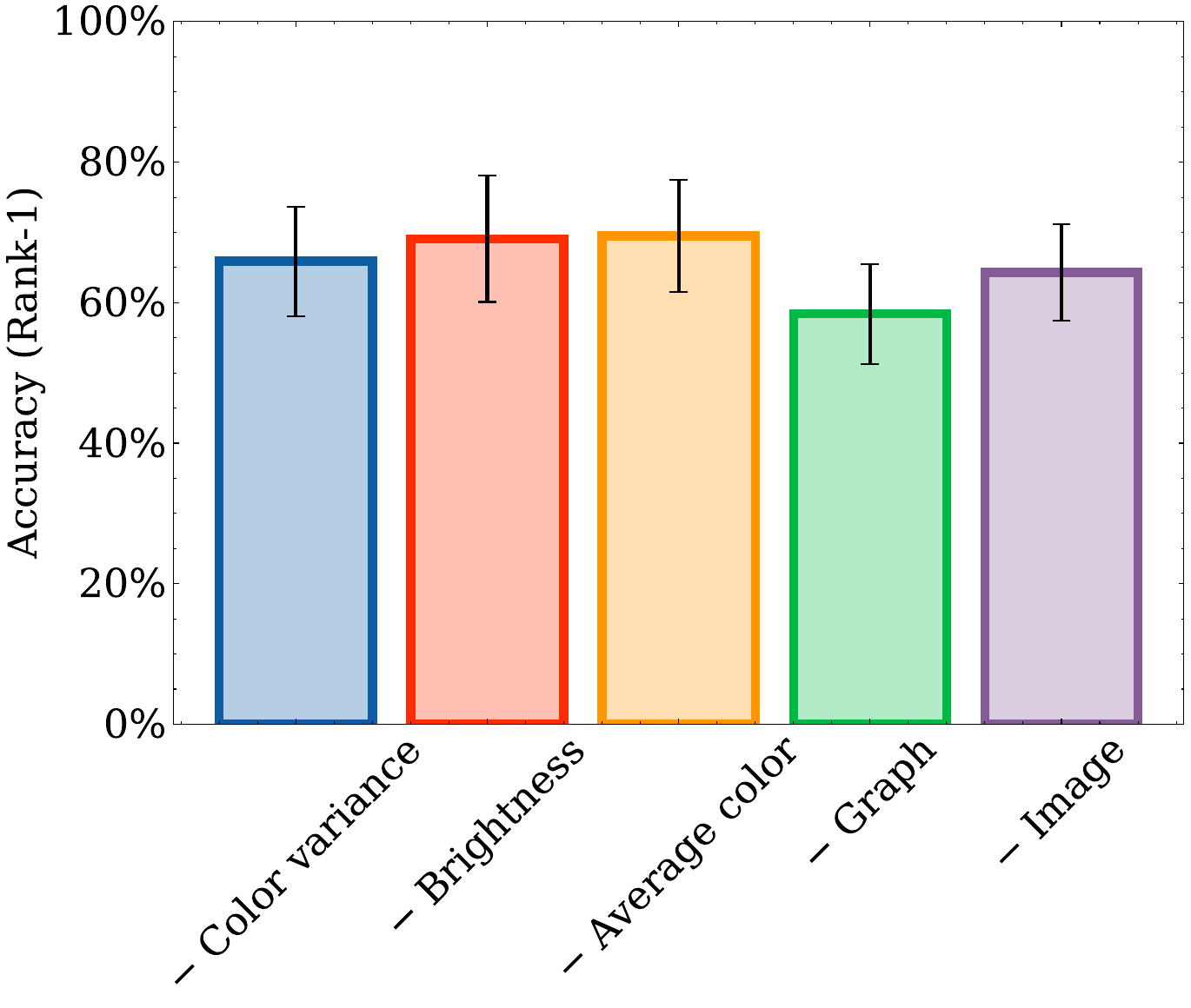}}
\caption{Ablation study: Rank-1 accuracy for our ensemble approach when removing one of the sub-model (the respective one noted on the x-axis).}
\label{fig:minus1}
\end{figure}

While we study the contribution of individual sub-model pairs in the main paper (Fig. 4), another way to characterize these contributions would be to study ensembles with only $4$ out of $5$ sub-models included. Thus, removing the most important sub-model shows the lowest resulting performance.
This is studied in Fig. \ref{fig:minus1}.

\newpage
While the graph shows that, as expected, the highest impact is achieved by the graph model, the differences are not significant, showing that which specific ensemble sub-models to use might not be the most important choice to make.

\subsection{Sub-model correlation}

We want to characterize how far the predictions of different sub-models differ from one another. Because we cannot simply use the Pearson Correlation for this, as the sub-model predictions are vectors, we define triplet correlations in Alg. \ref{alg:corr}.

The correlation between two functions is calculated via the likelihood that said functions provide the same order for the distances between three samples.
The results are normalized so that the correlation follows the usual range of $-1$ to $1$, with $0$ representing a random chance.
\newpage
\begin{algorithm}[h!]
\caption{Calculation of  $corr_{triplet}$}\label{alg:corr}
\begin{algorithmic}
\Require $f$,$g$, $x \in X$, $accuracy$
\State $count \gets 0$
\State $success \gets 0$

\While{$count < accuracy$}
\State $sample\; (A,B,C)\; from X$
\State $\Delta_f^{A,B}=\|f(A)-f(B)\|_2$
\State $\Delta_f^{A,C}=\|f(A)-f(C)\|_2$
\State $\Delta_g^{A,B}=\|g(A)-g(B)\|_2$
\State $\Delta_g^{A,C}=\|g(A)-g(C)\|_2$
\If{$\Delta_f^{A,B}<\Delta_f^{A,C} \And \Delta_g^{A,B}<\Delta_g^{A,C}$}
    \State $success \gets success +1$
\EndIf

\If{$\Delta_f^{A,B}>\Delta_f^{A,C} \And \Delta_g^{A,B}>\Delta_g^{A,C}$}
    \State $success \gets success +1$
\EndIf

\EndWhile

\State $corr_{triplet} \gets 2 \cdot \frac{success}{count}-1$
\end{algorithmic}
\end{algorithm}

\begin{figure}[h!]
\centerline{\includegraphics[width=0.8\columnwidth]{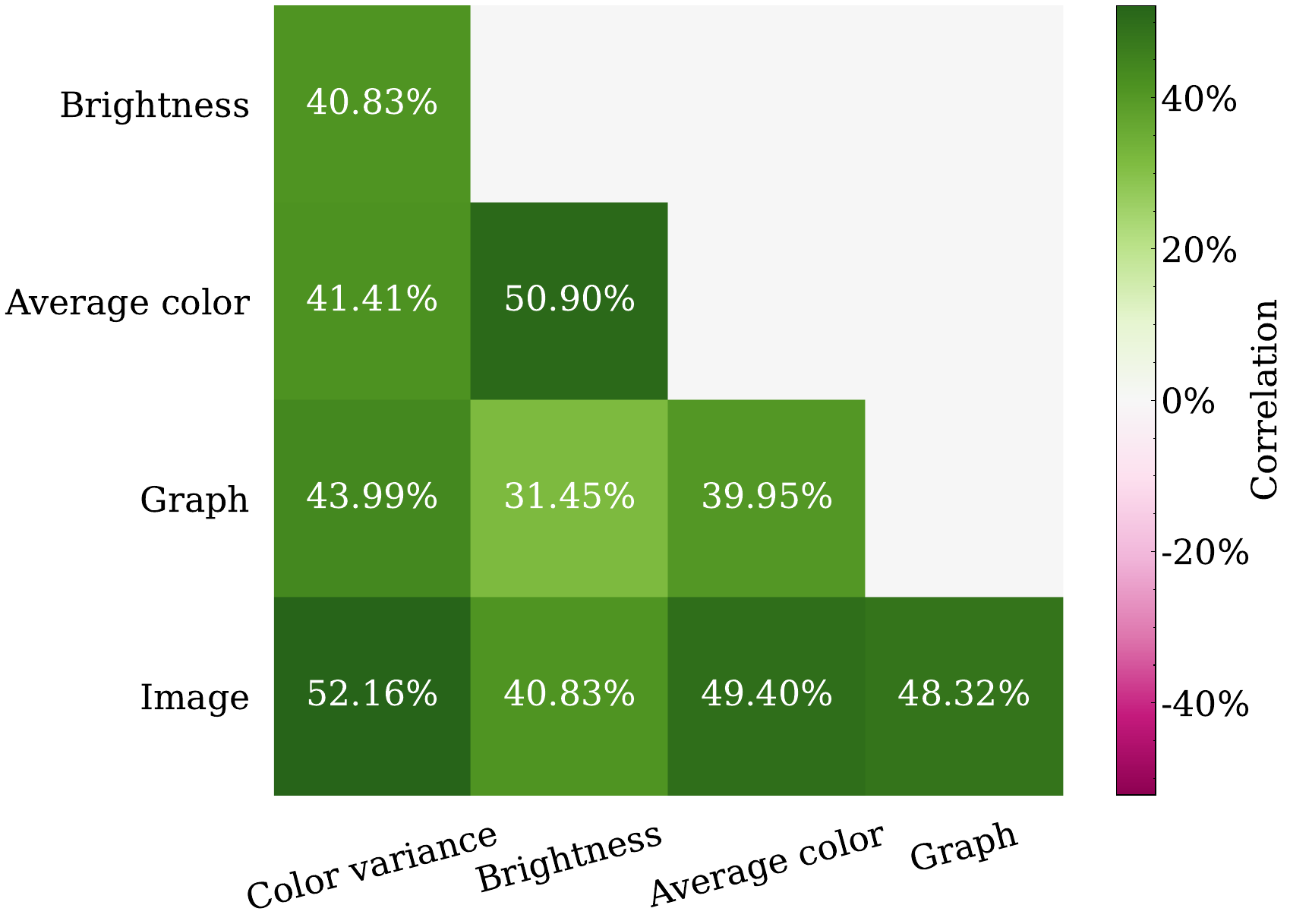}}
\caption{Correlation (as defined in Alg. \ref{alg:corr}) between sub-model predictions.}
\label{fig:corr}
\end{figure}

While some correlations between our individual sub-models can be perceived (see Fig. \ref{fig:corr}), there does not seem to be a significant difference between correlations nor an interesting pattern. This is especially significant when comparing this plot with Fig. \ref{fig:duals}.

\section{Supplementary Material -- Model Specifications}
In this subsection we include some hyperparameters that are interesting to reproduce our work, but where we did not find space for in the main paper

\subsection{Image based submodel}
Since the original images in the dataset are high-resolution images, they are resized to a dimension of $400\times230$ px.
The resolution of the original images varies, since they have been automatically cropped based on YOLO's bounding boxes, as is described in \mbox{\cite{rutinowski2021}}.
Their average resolution, however, is $1814\times1096$ px.

The transformation in our approach consists of six convolutional layers, each increasing the number of features by $1.5$ with a kernel size of $3$. 
After every other convolution, we employ a max-pooling operator with a kernel size of $2$ and add 3 dense layers to learn a final representation with $100$ dimensions.
We employ ReLU as an activation function, a learning rate of $0.001$ and a batch size of $256$.

\section{Supplementary Material -- Tabular Version of Main Paper Figures}

This section provides further tables that correspond to figures included in the main body of work.
They can provide interested readers with more details of the results and their standard deviations, and allow them to be further analyzed.

\begin{table*}[ht]
\renewcommand{\arraystretch}{1.5}
         \setlength\tabcolsep{7pt}
\centering
\caption{Correlations (see Alg. \ref{alg:corr}) of sub-model predictions. The correlations are not quite symmetrical due to the use of random samples.}
    \begin{tabular}{lrrrrr}
\hline
               Model & Color variance   & Brightness   & Average color   & Graph   & Image   \\
\hline
 Color variance & -                & $40.5\%$           & $40.9\%$              & $43.5\%$      & $52.0\%$      \\
 Brightness     & $40.8\%$               & -            & $51.3\%$              & $31.4\%$      & $41.1\%$      \\
 Average color  & $41.4\%$               & $50.9\%$           & -               & $39.4\%$      & $48.6\%$      \\
 Graph          & $44.0\%$               & $31.4\%$           & $39.9\%$              & -       & $47.5\%$      \\
 Image          & $52.2\%$               & $40.8\%$           & $49.4\%$              & $48.3\%$      & -       \\
\hline
\end{tabular}
\end{table*}

\begin{table*}[ht]
\renewcommand{\arraystretch}{1.5}
         \setlength\tabcolsep{8pt}
    \centering
\caption{Tabular version of Fig. 3., showing the ranked accuracy of the ensembles and the best performing sub-model on \cite{rutinowski_jerome_2021_6353714}.}

    \begin{tabular}{lrrr}
\hline
 Model             & Rank-1 Accuracy                 & Rank-6 Accuracy                & Rank-10 Accuracy               \\
\hline
 Concatenation    & $0.703 \pm 0.079$ & $0.92 \pm 0.032$  & $0.964 \pm 0.02$  \\
 Weighted Rank-1  & $0.675 \pm 0.095$ & $0.914 \pm 0.033$ & $0.952 \pm 0.012$ \\
 Weighted Triplet & $0.49 \pm 0.043$  & $0.731 \pm 0.046$ & $0.811 \pm 0.063$ \\
 Majority Vote    & $0.37 \pm 0.033$  & $0.584 \pm 0.02$  & $0.627 \pm 0.016$ \\
 NN Triplet       & $0.331 \pm 0.047$ & $0.663 \pm 0.053$ & $0.759 \pm 0.058$ \\
\hline
\end{tabular}
\end{table*}

\begin{table*}[ht]
\renewcommand{\arraystretch}{1.5}
         \setlength\tabcolsep{7pt}
    \centering
    \caption{Tabular version of Fig. 4, showing the improvement in Rank-1 accuracy of pairs of models over their expectation.}
\begin{tabular}{lrrrrr}
\hline
                Model & Color variance   & Brightness   & Average color   & Graph    & Image    \\
\hline
 Color variance & -                & $5.7\%$            & $8.7\%$               & $6.2\%$       & $7.2\%$       \\
 Brightness     & $5.7\%$                & -            & $12.6\%$              & $1.6\%$       & $2.4\%$       \\
 Average color  & $8.7\%$                & $12.6\%$           & -               & $-1.4\%$      & $1.2\%$       \\
 Graph          & $6.2\%$                & $1.6\%$            & $-1.4\%$              & -      & $9.1\%$       \\
 Image          & $7.2\%$                & $2.4\%$            & $1.2\%$               & $9.1\%$       & -       \\
\hline
\end{tabular}
\end{table*}

\begin{table*}[ht]
\renewcommand{\arraystretch}{1.5}
         \setlength\tabcolsep{8pt}
    \centering
\caption{Tabular version of Fig. 10, showing the ranked accuracy of the ensembles and the best performing sub-model on \cite{rutinowski_jerome_2022_7386956}.}

\begin{tabular}{lrrr}
\hline
 Model             & Rank-1 Accuracy                & Rank-6 Accuracy                & Rank-10 Accuracy               \\
\hline
 Weighted Rank-1  & $0.777 \pm 0.054$ & $0.975 \pm 0.01$  & $0.992 \pm 0.007$ \\
 Concatenation    & $0.698 \pm 0.069$ & $0.921 \pm 0.032$ & $0.958 \pm 0.011$ \\
 Majority Vote    & $0.598 \pm 0.092$ & $0.791 \pm 0.06$  & $0.813 \pm 0.06$  \\
 Weighted Triplet & $0.349 \pm 0.04$  & $0.802 \pm 0.043$ & $0.902 \pm 0.028$ \\
 NN Triplet       & $0.349 \pm 0.065$ & $0.77 \pm 0.036$  & $0.866 \pm 0.037$ \\
\hline
\end{tabular}
\end{table*}

\end{document}